\documentclass[10pt,twocolumn,letterpaper]{article}

\usepackage{caption} 
\usepackage{authblk}
\usepackage{wacv}
\usepackage{times}
\usepackage{epsfig}
\usepackage{graphicx}
\usepackage{amsmath}
\usepackage{amssymb}
\usepackage{booktabs}
\usepackage{enumitem}
\usepackage{adjustbox}
\usepackage{placeins}
\usepackage{subfiles}

\usepackage[utf8]{inputenc}

\wacvalgorithmstrack   
\wacvfinalcopy 

\usepackage[pagebackref=true,breaklinks=true,colorlinks,bookmarks=false]{hyperref}

\begin{document}

\title{Anticipating Next Active Objects \\ for Egocentric Videos}

\author[1,4]{Sanket Thakur}
\author[2, 1]{Cigdem Beyan}
\author[1]{Pietro Morerio}
\author[3, 1]{Vittorio Murino}
\author[1]{Alessio {Del Bue}}
\affil[1]{Pattern Analysis and Computer Vision (PAVIS), Istituto Italiano di Tecnologia (IIT)}
\affil[2]{Department of Information Engineering and Computer Science (DISI), University of Trento, Italy}
\affil[3]{Department of Computer Science, University of Verona, Italy}
\affil[4]{Department of Electrical, Electronics and Telecommunication Engineering and Naval Architecture (DITEN), University of Genoa, Italy}
\maketitle
\thispagestyle{empty}

\begin{abstract}
   This paper addresses the problem of anticipating the next-active-object location in the future, for a given egocentric video clip where the contact might happen, before any action takes place. The problem is considerably hard, as we aim at estimating the position of such objects in a scenario where the observed clip and the action segment are separated by the so-called ``time to contact'' (TTC) segment. Many methods have been proposed to anticipate the action of a person based on previous hand movements and interactions with the surroundings. However, there have been no attempts to investigate the next possible interactable object, and its future location with respect to the first-person's motion and the field-of-view drift during the TTC window. We define this as the task of Anticipating the Next ACTive Object (ANACTO). To this end, we propose a transformer-based self-attention framework to identify and locate the next-active-object in an egocentric clip.
    We benchmark our method on three datasets: EpicKitchens-100, EGTEA+ and Ego4D. We also provide annotations for the first two datasets. Our approach performs best compared to relevant baseline methods. We also conduct ablation studies to understand the effectiveness of the proposed and baseline methods on varying conditions. Code and ANACTO task annotations will be made available upon paper acceptance.
\end{abstract}

\section{Introduction}

\begin{figure}[t!]
\begin{center}
    \includegraphics[width=\linewidth]{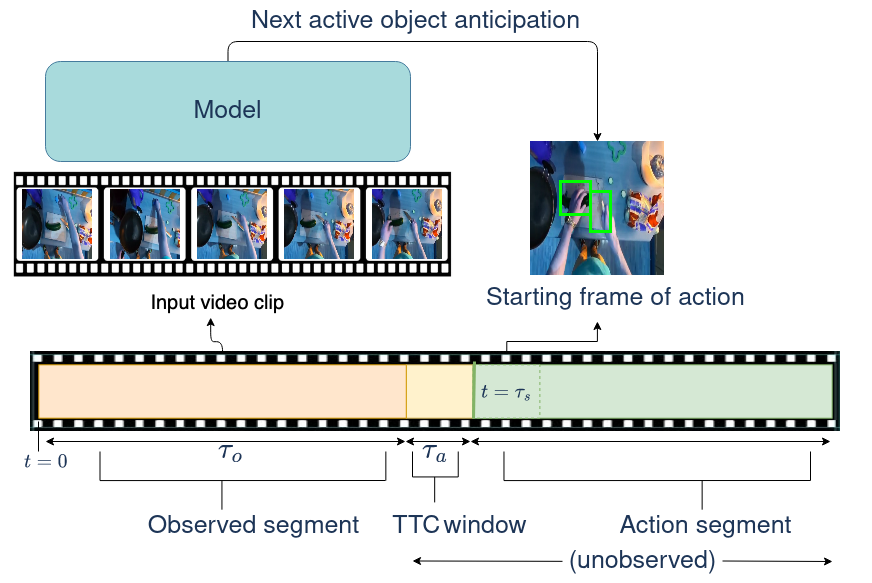}
\end{center}
\caption{The goal of our work is to anticipate the next-active-object, i.e. to localize the object that the person will interact with in the first frame of an action segment, based on the evidence of video clip of length $\tau_o$, located $\tau_a$ seconds (anticipation time) before the beginning of an action segment at time-step $t = \tau_s$.}
\centering
\label{fig:teaser}
\vspace{-15pt}
\end{figure}

The widespread use of wearable cameras prompted the design of egocentric (first-person) systems that can readily support and help humans in their daily activities, by augmenting their abilities \cite{youdoilearn_1,youdoilearn,fpv}. In order to assist users, a fundamental problem is to predict, forecast and even anticipate what the person will do in the next few second(s). Among all the possible tasks, one of the most relevant is to understand from an egocentric video stream, which object a user will interact with or manipulate in the near future. Besides, it is not just enough to localize the next-active-object (\emph{NAO}) but also to model the motion and Field-of-View (FoV) drift till the contact with the object actually happens. Solving this task can help to gain more understanding about the future activity of the person as well as the usage of the objects. However, compared to other tasks performed with egocentric videos, anticipating interactable objects is notably challenging since humans interact with the environment based on their final goals and the responses they get from the environment. On the other hand, performing this task is useful, for example, by doing so 
a robot can prevent a collision between object(s) and human(s) in a warehouse by analysing the past observation and estimating the future point of contact or provide support in human-robot interactions for instance in factories where objects are also moving to anticipate the contact location based on robot movement \textit{wrt} objects.

In this paper, we call this task  \textit{``Anticipating Next ACTive Object''} (ANACTO) by following the nomenclature of the most recent literature \cite{ego4d}. In \cite{ego4d}, \emph{NAO} is defined for the object which is identified in the \emph{last observed frame}. Instead, our ANACTO task further expands this definition to formulate the motion and FoV drift of the interactant to anticipate the \emph{NAO} at its contact point. According to \cite{ADL}, active objects are those which are in contact (usually with hands) with the first-person. However, our work focuses on the localization of the \emph{NAO} after a certain time at its contact point \emph{before} any interaction(s) begin, as shown in Figure \ref{fig:teaser}. We have past evidence from observed video clip segment of length $\tau_o$, which precedes the actual action by a \textit{time to contact window} $\tau_a$. We define ANACTO as the task of predicting the bounding box of \emph{NAO} involved in the action in its starting frame(s) at its contact point ($t=0$).  Notice that, ANACTO task refers to not only detect/localize the NAO in the last observed frame (which is the case for Ego4D's Short-term anticipation (STA)) but also anticipating the final location of NAO at which the contact/interaction actually happens even in much later upcoming frames. 
Instead, Ego4D STA does not aim to identify the final interaction with the object.
In Ego4D STA, it is assumed that object are static because of the fact that the last observed frame is considered only.

We propose to address the ANACTO task by exploring the combination of object-centered and human-centered cues while leveraging the self-attention mechanism of vision transformers (VIT) \cite{vit}. In detail, the proposed method analyzes RGB frames to gain an understanding of hand's position and their motion without explicitly using hand information. At the same time, it exploits an object detector to include spatial positioning of objects in the observed clip. Since ego-actions are mainly characterized as the interaction between the user's hands and objects in the scene, we claim that VIT's self-attention is a good candidate for capturing such relationships, both on frame-level and across frames. Indeed, the correctness of this claim is shown by quantitative (which also includes comparisons with several relevant methods) and qualitative analysis.
The main contributions of this work are the following: \vspace{-0.2cm}
\begin{itemize} [leftmargin=*]
 \setlength\itemsep{0em}
\item A new task called Anticipating the Next ACTive Object (ANACTO) in egocentric videos is introduced.
\item A novel method to address ANACTO, which is based on vision transformers, encoding the interactions between the first-person and the objects, and accounting for the time to contact window, is proposed.
\item Existing action anticipation state-of-the-art (SOTA) methods are extended to perform ANACTO task.
\item Our method as well as the SOTA are benchmarked on EpicKitchens-100 \cite{ek100} (EK-100),  EGTEA+ \cite{egtea} and Ego4D \cite{ego4d} datasets. The performance comparisons among all methods prove the effectiveness of the proposed method in all cases. For the EK-100 and EGTEA+ datasets, we also provide annotations for the ANACTO task.
\end{itemize}
\section{Related Work} \label{sec:relatedwork}

We first review studies on egocentric action anticipation, since our problem follows a similar approach. Yet, instead of action classification, we focus on regressing the location of \emph{NAO}. Then, we review the definition of ``active'' objects, which are also closely related to \emph{NAO}, and then investigate the existing works on it. \\
\\
\noindent
\textbf{Action Anticipation in Egocentric Videos.}
Action anticipation is the task of predicting future actions \emph{before} they occur.
The anticipation problem has been well-explored for various actions from \emph{third person videos} \cite{kitani2012activity,lan2014hierarchical,huang2014action,jain2015car,felsen2017will,gao2017red,Farha_2018_CVPR,lan2014hierarchical,Vondrick_2016_CVPR,Rodriguez_2018_ECCV_Workshops}.
Instead, its application in \emph{first-person videos}, which is formalized in \cite{ek55}, has only recently gained popularity \cite{liu2019forecasting,rulstm,miech2019leveraging,tpami_contact} due to its applicability on wearable computing platforms \cite{avt}.
We discuss works that are closely related to our anticipation task such that perform short-term (i.e., ``recent'', see \cite{Sener2020} for its definition) egocentric action anticipation since we have evaluation protocols and datasets in common.

Lui et al. \cite{liu2019forecasting} define the egocentric action anticipation problem in terms of human-object interaction forecasting, in which the hand movement is used as a feature representation to predict the egocentric hand motion, interaction hotspots and the future action. Dessalene et al. \cite{tpami_contact} perform hand-object contact and activity modeling to anticipate partially observed and/or near future action. For hand-object contact modeling, the short-term dynamics is learned with 3D Convolutions. The localization of boundaries between the hands and objects in contact is performed by applying segmentation through a U-Net \cite{ronneberger2015u}. The activity modeling stage embeds the output of contact modeling through Graph Convolutional Network (GCN) layers \cite{kipf2016semi} and then fed to an LSTM, which is followed by a fully-connected layer to make action predictions. On the other side, there exist methods relying on the aggregation of the information from the past frames in an observed video clip \cite{rulstm,miech2019leveraging}.
For example, \cite{rulstm} propose RU-LSTM, a method  composed of a ``rolling'' LSTM (R-LSTM) encoding the past observations, and the ``unrolling'' LSTM (U-LSTM) taking over the current hidden and cell states of the R-LSTM and producing hypotheses of future actions. 
Differently, the model in \cite{miech2019leveraging} uses a predictive model (a CNN) and a transitional model (a CNN pre-trained on action recognition). The predictive model directly anticipates the future action while the transitional model is constrained to the output of the currently happening action that is later on used to anticipate the future actions.
Recently, \cite{avt} presented an architecture based on transformers to encode the data performed by the backbone and predict the future actions performed by the head network. \cite{avt} achieves superior results compared to \cite{rulstm} and shows the better performance of transformer backbone with respect to using many other backbones such as TSN \cite{TSN2016ECCV} and Faster R-CNN \cite{fastercnn}. Compared to \cite{avt}, our transformer based architecture additionally aims to exploit the object-centric features with spatial and temporal attention along with two losses introduced to model past observation and learn about active object(s) to anticipate \emph{NAO} at its contact point using an autoregressive decoder.

Since our ANACTO task is novel, to obtain relevant baselines to compare with, we have modified several action anticipation SOTA tested on egocentric videos \cite{liu2019forecasting,rulstm,avt} and tested on third-person videos \cite{TSN2016ECCV} (we include \cite{TSN2016ECCV} due to its promising results demonstrated in \cite{avt} for egocentric settings).
For the baselines \cite{liu2019forecasting,TSN2016ECCV}, we append our decoder (see \ref{section:decoder} for its definition) to aggregate the frame level information gathered from their backbone in order to perform the ANACTO task.
In terms of encoder design, as we propose a transformer based architecture, our method differs from \cite{liu2019forecasting,rulstm,TSN2016ECCV} which are based on I3D-Res50 \cite{i3d}, LSTMs \cite{yu2019review}, and Temporal Segment Networks, (i.e., Spatial and Temporal ConvNets), respectively. \\
\\
\noindent
\textbf{Active Objects.} For the first time, \cite{ADL} defined \emph{active} and \emph{passive} objects in an egocentric setup. Their method is based on the appearance differences among the objects (e.g., an opened fridge is an active object which looks different from a closed fridge called a passive object), and the location of the active object (i.e., active objects tend to appear close to the center of an egocentric image). By definition, active objects are those, which are currently involved in an interaction, e.g., being touched by humans, whilst, the passive objects are the background objects that the human agent is not in an interaction with, e.g., not manipulating them \cite{ADL}. 
Dessalene et al. \cite{tpami_contact} adapted these definitions to describe \emph{NAO}, which stands for the object that will be contacted with a hand. Their method requires the visibility of the \emph{NAO} and the existence of the hands in the current frames. It was also only tested when some specific action classes (take, move, cut and open) were considered. Instead, our method processes the frames independent to the hand(s) visibility or presence in the current frames. Importantly, we do not specifically restrict the possible (inter)actions between the human and the objects, i.e., we use all the verb classes supplied by the benchmark datasets. \cite{jiang_NAO} also explored \emph{NAO} prediction using cues from visual attention and hand position, but by only using a single frame for the prediction. That approach \cite{jiang_NAO} is not able to differentiate between the past or future active object, since it does not account for the temporal information acquired by the videos. Furnari et al. \cite{furnari2017next} also explored the \emph{NAO problem} by taking into account the active/passive objects definition of \cite{ADL}. Their method \cite{furnari2017next} uses an object tracker to extract the object trajectories for a small video clip till the last frame precedes an action. This trajectory is later used to classify whether a given object is going to be active or passive in next frame. Such methodology \cite{furnari2017next} is restricted to predicting the \textit{immediate NAO} instead of predicting the location of the active objects in several future frames as our proposed method can do. Moreover, it requires an observation time which is till the penultimate frame of an action segment, which is unpredictable in real-life implementations.
Very recently, Liu et al. \cite{Liu_2022_CVPR}, proposed a similar setup but to forecast hand trajectories for interaction hotspots on next-active-objects, i.e. confining to human hands interactions. Instead, our setup is more generic, e.g., can include interactions of robot.

\section{ANACTO in Egocentric Videos} \label{naosetup}
In this section, we first formalize the ANACTO problem, and then we provide details about the proposed model. Specifically, let $V$ be a given video clip, we split the video clip into three sequential parts: the observed segment of length $\tau_o$, the time to contact (TTC) window of length $\tau_a$ and a given action segment which starts at timestep $t = \tau_s$.
The goal is to localize \emph{NAO} at the beginning of an action segment at timestep $t = \tau_s$ where the contact happens, using $\tau_o$ length observed video clip $\tau_a$ seconds \emph{before} the beginning of the action segment involving \emph{NAO} (see Fig. \ref{fig:teaser}). In other words, ANACTO is a combination of two tasks merged into one: 
(1) identifying \emph{NAO} from past observed segment, and
(2) by using the past observation(s), modelling the motion of a person to estimate \emph{NAO}'s location after the TTC window where actual contact happens.
Notice that, this definition assumes that for every action to be performed, a person interacts with an object either with their hands or with a tool such that the object becomes active at the starting point of the action. Therefore, our problem description is not bounded with ``hand''-object interactions only, consequently our approach does not include/require the detection of hands (e.g. the physical interactions can be performed by a tool as well).

\begin{figure*}[t!]
\centering
\includegraphics[width=0.85\linewidth, height=0.4\linewidth]{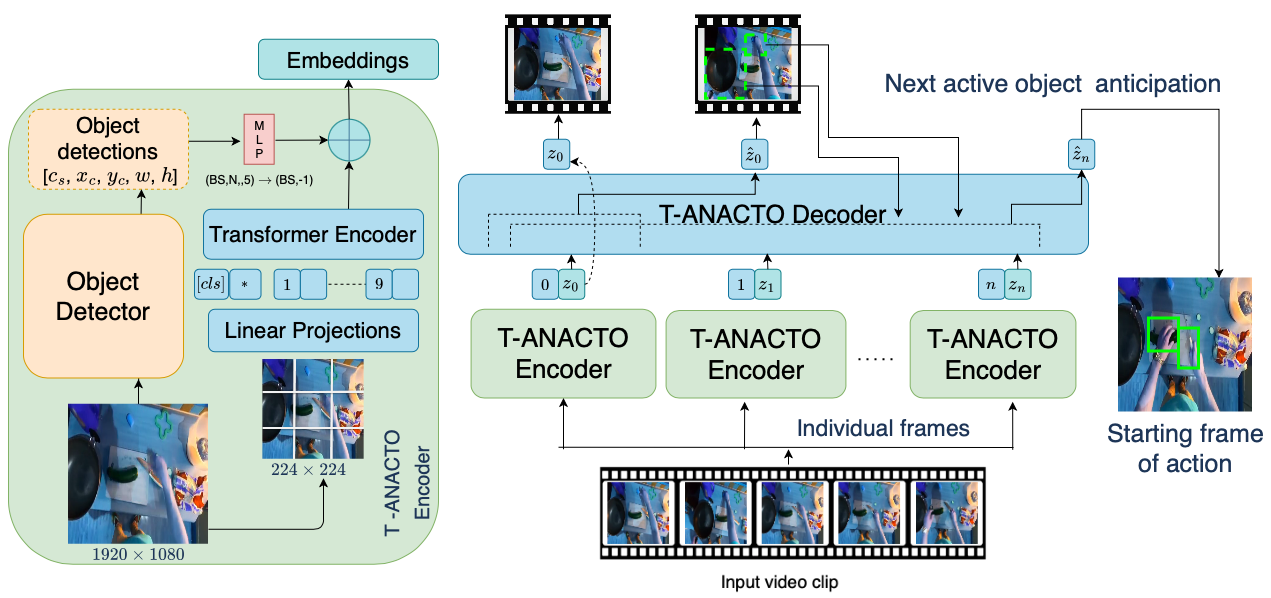}
\caption{Our T-ANACTO model is an encoder-decoder architecture. Its encoder is composed of an \emph{object detector} and a \emph{Vision Transformer} \cite{vit}. The object detector \cite{fastercnn} takes an input frame (e.g., size of 1920$\times$1080) and predicts the location of objects in terms of bounding boxes ($x$, $y$, $w$, $h$) and detection confidence scores ($c$). The input of VIT are the frame(s), first resized to, 224$\times$224 and then divided into the patches (16$\times$16). The object detections ($x$, $y$, $w$, $h$) are also converted to match the scaled size of the frame (i.e., 224$\times$224), reshaped, and are then passed through a MLP to convert it into the same dimension as the embeddings from the transformer encoder, which are later concatenated together to be given to the decoder. There exist a linear layer between the decoder and the T-ANACTO encoder, which adjusts the feature dimensions to be fed to the transformer decoder. Transformer decoder uses temporal aggregation to predict the next active object. For each frame, the decoder aggregate the features from the encoder for current and past frames along with the embeddings of last predicted active objects and then predicts the next active object for the  future frames.
} 
\centering
\label{fig:method}
\vspace{-15pt}
\end{figure*}

\subsection{Proposed Method: T-ANACTO}
\label{sec:proposedMethod}
We propose a method, which regresses the location of \emph{NAO} from egocentric videos by analyzing the past video frames, and incorporating object detections for the input frames. Object detections refer to the location of the object bounding box ($x_c$, $y_c$, $w$, $h$), and a confidence score ($c_s$) produced by the detector. Fig. \ref{fig:method} illustrates the proposed method. 

The proposed method (called T-ANACTO stands for Transformer-based Anticipating Next ACTive Object) leverages the self-attention mechanism of VIT to construct a \textit{encoder} network that operates on individual frames or short clips, followed by a transformer \textit{decoder}. 
The \emph{T-ANACTO encoder} consists of a vision transformer (VIT) \cite{vit} and an object detector \cite{fastercnn} which are used to extract the feature embeddings from each video frame. 
Our decoder is inspired from \cite{avt}, such that we exploit its \emph{causal} structure - to tackle a predictive task based on past observations and make it autoregressive for an egocentric setting. This model choice was supported by the fact that Transformer-based end-to-end attention methods are efficient not only in recognizing actions in given-video segments, but also in predictive video modelling. There also exist promising results in anticipation and object detection-based tasks on static images \cite{detr,avt,fang_yolos,kim2021hotr,farinella_egosurvey}.
The \textit{T-ANACTO decoder} aggregates the information acquired in temporal dimension to collectively understand the first-person's movements with a final goal of predicting the location of the \emph{NAO}. Herein, we also introduce 2 losses to enforce the model to attend to past active objects to predict for NAO in future frames based on previous observations.

\subsection{T-ANACTO Encoder}
The encoder of our model consists of an object detector \cite{fastercnn} (called as object detection head, ${H_o}$) combined with a VIT \cite{vit} (i.e., a video backbone). The object detector identifies the positions of the objects, while VIT analyzes a RGB video frame to understand the context. Different from \cite{avt}, we demonstrate the importance of object-centric features with temporal attention along with two losses introduced to model past observation and anticipate future contact point, described in detail below.

\begin{figure}[t!]
\begin{center}
    \includegraphics[width=\columnwidth]{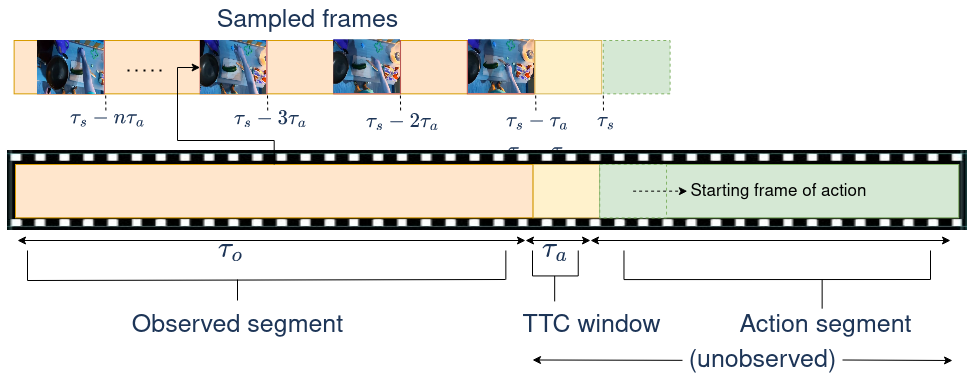}
\end{center}
\caption{The observed video segment of length $\tau_o$ is sampled at a frame rate equal to the TTC time (shown as $\tau_a$) to maintain consistency in (1) the frame interval of sampled frames and (2) between the last observed frame and the starting frame of the action segment, which starts at $t = \tau_s$. 
}
\centering
\label{fig:sampling}
\vspace{-10pt}
\end{figure}

Given a video clip $V$ = $\{X_1, X_2, \dots X_T\}$ with $T$ frames, where ${X_t}$ is the RGB image at time step $t$ and an action segment, we trim the video clip into: (1) observed segment length of $\tau_o$, (2) TTC window ($\tau_a$), before the beginning of the action segment at $t = \tau_s$. Frames from the observed segment are then sampled at a frame rate which is equal to $\tau_a$ to maintain consistency between frame intervals as described in Fig. \ref{fig:sampling}. Each frame extracted from the observed segment is an input of an individual T-ANACTO Encoder. Our object detection head ${H_o}$ follows a Faster R-CNN \cite{fastercnn} architecture and consists of a region proposal network and a regression head. It takes as input each RGB frame $X_t$ and generate bounding boxes $b_{i,t}$ $\in$ $\mathbb{R}^4$ with corresponding confidence score ${cs}_{i,t}$ $\in$ $(0, 1)$ such that:
\begin{equation}
    b_{i,t},{cs}_{i,t} = H_o(X_t), \:\:\: i \in \{1, \ldots N\},
    \vspace{-3pt}
\end{equation}
where $N$ is the total categories of objects for a dataset. For a category, detections with the highest confidence score are used.

The object detections are performed for the original size of the image frames, $X_t$ (e.g. $1920 \times 1080$) and then the bounding boxes are scaled to match the resized image size, $X^r_t$ of $224 \times 224$ to match with the input size of VIT \cite{vit}. The detections are then reshaped $(BS, N, 5) \xrightarrow{} (BS, -1)$ to be passed through an MLP,  ${f_{MLP}}$ to convert them to the same dimensions as the T-ANACTO encoder's output.

For our video backbone $V_b$, we adopt ViT-B/16 using $224 \times 224$ images, where ${X^r_t}$ is an image at a time $t$.
We split each input frame into $16 \times 16$ non-overlapping patches, which are later flattened into a 256-dimensional vector. The vector representation is then projected to a 768-dimensional vector to be used as the input for our transformer encoder. The feature dimensions are kept constant throughout the encoder. 
We also append a learnable \textit{[cls]} token in the patch features, which can later be used to identify  active object(s) label in the current frame, if any. All the other patches are also allocated a spatial positional embedding with their patch embedding. The resulting patch embeddings are then passed through a standard VIT Encoder with pre-norm. Finally, the feature representations learnt for each frame from the visual backbone are concatenated with the detections obtained from the object detection head as follow: 
\begin{equation}
    z_t = V_b(X^r_t) + f_{MLP}(H_o(X_t)).
    \vspace{-3pt}
\end{equation}
In the end, we add a temporal position encoding to the extracted features from the T-ANACTO encoder for each frame, which are further given to the decoder network.

\subsection{T-ANACTO Decoder} \label{section:decoder}
We argue that the past observations can provide a lot of context to produce hypothesis regarding the \emph{NAO}. Therefore, for the decoder network, we take inspiration from \cite{avt}, and extending it to make it autoregressive at each step, to aggregate the features of the past frames and exploit the last predicted active object location which allows us to perform ANACTO.

The decoder network $D$ is designed to produce attentive features corresponding to the future frames: $\hat{z_1}, \dots , \hat{z_t}$ to anticipate the location of the \emph{NAO} for each input frame as: 
   $ \hat{z_t} = \textit{D}(z_0, \dots , z_{t}; \hat{h_0}, \dots , \hat{h_{t-1}}) $ (see also Fig. \ref{fig:method}).
Here $\hat{z_t}$ is the predicted features of the \emph{future frame} at t+1 obtained after attending to all other encoded features belonging to the frames before t+1 (i.e., $z_0, z_1 .. z_t$). At each frame, the decoder takes the previously predicted active object location $\hat{h_t}$ in previous frames along with RGB features to estimate the next-active-object position, $\hat{y_t}$ in future frames. Both these features are concatenated together and are then fed to the next step. 
This helps in aggregating features of the past frames and understanding the intention and final goal of the first-person, which is defined by the action segment ground-truth label. These features are passed through multiple decoder layers, each consisting of masked multi-head attention, LayerNorm ($LN$), and a multi-layer perceptron (MLP) as in \cite{gpt2}. The final output is then passed through another $LN$ to obtain the final embeddings. For each decoder output $\hat{z_t}$, it is used to regress the \emph{NAO} in the corresponding frame at t+1. The predicted features are then fed to a linear layer $\theta$, to regress the bounding box coordinates $\hat{y_t}$ $\in$ $\mathbb{R}^8$, \textit{i.e.} $\hat{y_t}$ = $\theta(\hat{z_t})$. The final prediction $y_t$ represents the model's output at each frame.

\subsection{Loss Calculation}
To train T-ANACTO, we sample a clip preceding each labeled action segment in a given dataset, ending $\tau_a$ seconds before the start of the action. The clip is then sampled with the same frame rate as $\tau_a$ seconds to remain consistent with frame intervals as described in Fig. \ref{fig:sampling}. The sampled frames are then passed through our T-ANACTO model and train the network in a supervised manner with three loss functions, described as follows. 

${L}_{feat}$ defined in Eq. \ref{eq_lfeat} aims at leveraging the predictive structure of the model by supervising the future frame features predicted by the decoder to match the true future frame features that are extracted as embeddings from the encoder. 
\begin{equation}
    \mathcal{L}_{feat} = \sum_{t=0}^{N}{||\hat{z}_t - z_{t+1}||}^2_2,
    \label{eq_lfeat}
    \vspace{-5pt}
\end{equation} where $N$ is the number of frames in training.
It is to be noted that our model does not need the presence of hand or any active object to be present in the observed segment. However, any active object found in the observed segment provides additional supervision using ${L}_{cao}$, stands for current active object loss, is a Mean Squared Error (MSE) Loss used for the prediction of active objects \emph{in the observed segment of the video clip}. In addition, $\mathcal{L}_{nao}$, stands for the next-active-object loss, forces the model to identify the location of the \emph{NAO} \emph{at the start of an action}. . It is supported by, $\mathcal{L}_{cao}$ which helps T-ANACTO to identify and keep track of active object(s) found at the end of the observed video segment.
\begin{equation}
    \mathcal{L}_{cao} = \sum_{t=0}^{N-1}||y_t - \hat{y_t}||^2,
    \mathcal{L}_{nao} = ||y_n - \hat{y_n}||^2,
    \label{eq_lcao}
    \vspace{-5pt}
\end{equation}
where $y_t$ $\in$ $\mathbb{R}^8$ and $\hat{y_t}$ $\in$ $\mathbb{R}^8$ are the ground-truth and predicted bounding boxes for active objects in \emph{the current frame}, respectively. Whereas $y_n$ $\in$ $\mathbb{R}^8$ and $\hat{y_n}$ $\in$ $\mathbb{R}^8$ are the ground-truth and predicted bounding box for \emph{NAO} in \emph{the starting frame} of an action after $\tau_a sec$, respectively. The final loss is a linear combination of the aforementioned three losses: 
\begin{equation}
    \mathcal{L} = \mathcal{L}_{feat} + \lambda_1\mathcal{L}_{cao} + \lambda_2\mathcal{L}_{nao},
    \label{eq:final}
\end{equation}
where $\lambda_1$, $\lambda_2$ are fixed weights. 

\section{Experimental Analysis}
The experimental analyses were conducted on three major egocentric video datasets, described in Sec. \ref{sec:dataset}. As this is the first time ANACTO task is being benchmarked, there is no existing method performing it. Therefore, we adapted the SOTA action anticipation methods to perform comparisons in Sec. \ref{baselines}. We described the implementation details of T-ANACTO in Sec. \ref{impDet}.
\subsection{Datasets}
\label{sec:dataset}
\noindent
\textbf{EK-100 \cite{ek100}.} Consists of about 100 hours of recordings with over 20M frames comprising daily activities in kitchens, recorded with 37 participants. It includes 90K action segments, labeled with 97 verbs and 300 nouns (i.e. manipulated objects). It supplies the annotations regarding the hand and object interactions, which are used for ANACTO. In detail, the aforementioned annotations are in terms of the prediction results of a hand-object interaction detector \cite{Shan20}, which provides the hand location, side, contact state, and a bounding box surrounding the object that the hand is in contact. Such detector \cite{Shan20} was trained on EK-55 \cite{ek55}, EGTEA \cite{egtea} and CharadesEgo \cite{charadesego} datasets, and applied on EK-55 \cite{ek55} dataset to annotate it with respect to the hand-object interactions.
We use the following annotations: the locations of both hands (i.e., the bounding boxes  $b \in$ $\mathbb{R}^8$), and the locations of the objects along with the contact state information at each frame of each video and, then curate the final ground-truth data for ANACTO problem. It is important to mention that the videos in this dataset were collected with different frame rates. In order to apply the methods: \cite{rulstm,TSN2016ECCV,liu2019forecasting} requiring frame rates fixed to 30 frame per second, we converted each video to this constant frame rate, thus the annotations regarding the hand locations and active objects' locations are also interpolated accordingly. \\

\noindent
\textbf{EGTEA+ \cite{egtea}.}  
Includes 28 hours of videos containing 106 action categories, which corresponds to 2.4M frames. There exist 10325 action segments associated to 19 verbs and 53 nouns (i.e., objects) that were recorded with 32 participants. It is important to notice that yet there exist no publicly available source supplying annotations needed to perform ANACTO for EGTEA+.
Therefore, we created the hand-object interaction annotations following the annotation pipeline of EK-100 \cite{ek100} dataset. These include: the hand locations (bounding boxes \textit{b} $\in$ $\mathbb{R}^8$ and the corresponding detection confidence scores) at each frame, the active object locations and their contact state. First, all the videos are converted to a constant frame rate of 30 fps. Then, each frame is fed to the hand-object interaction detector model from \cite{Shan20}. The hand and object threshold is kept at 0.5 to produce better qualitative results, which is also the same when extracting the annotations for EpicKitcen-100 dataset \cite{ek100}. Additionally, we provide annotations for the videos with original frame rate for its original frame size. \\

\noindent
\textbf{Ego4D \cite{ego4d}.}  
This is the largest first-person dataset recently released. The dataset is split into 5 different categories, each focusing on a different task, combining for a total of 3,670 hours of egocentric videos across 74 locations. 
For this task, we focus on the forecasting split, containing ~1000 videos for a total of 960 hours, annotated at 30 fps for the short term interaction anticipation task. The annotations are for the \emph{NAO} in the \emph{last observed frame}. 

\subsection{Baseline Methods}
\label{baselines}
We compare T-ANACTO with SOTA action anticipation methods, namely AVT \cite{avt}, RULSTM \cite{rulstm}, Liu et al. \cite{liu2019forecasting} and TSN  \cite{TSN2016ECCV}. For RULSTM \cite{rulstm}, we used pre-extracted RGB, flow and object features as in their paper, for EK-100 and EGTEA+. For Ego4D, we computed the flow and RGB features by following the same TSN model mentioned in \cite{rulstm}, which were then fed as the inputs to the RULSTM model. We also tested individual modalities with TSN \cite{TSN2016ECCV} (ResNet101) for RGB frames and RULSTM-object centric path for object modality.
Moreover, we used object detections as well as their confidence score from the object detector \cite{fastercnn} to be used as object features in RULSTM(fusion) and RULSTM(obj). 
We modified and re-trained all these aforementioned methods in order to perform ANACTO task. We explored these methods (noticed that they were used for action anticipation in egocentric videos, a.k.a. a \emph{classification task}) because our problem formulation is very much related to action anticipations, and we claim that these methods can provide effective learning for ANACTO \emph{regression task} by modelling past motion. For each model, we replace the last classification layer with a regression layer to predict the bounding boxes $\hat{y_n}$ $\in$ $\mathbb{R}^8$ regarding the next active object. Since TSN \cite{TSN2016ECCV} method processes individual frames and not a video clip, for the corresponding experiments, we appended the whole T-ANACTO decoder layer to the TSN \cite{TSN2016ECCV} method allowing the aggregation of information from all frames (i.e., tuning the task from frame-level processing to video processing). Throughout this paper, we refer to these methods as \emph{baselines}.

\subsection{Implementation Details of T-ANACTO}
\label{impDet}
T-ANACTO was trained with an SGD optimizer for 50 epochs with a learning rate of $1e-5$. Recall that a linear layer exist after the output of the decoder to regress the bounding box coordinates $\hat{y_t}$ $\in$ $\mathbb{R}^8$ (here the results are in $\mathbb{R}^8$, confining to a single active object for each hand. We kept the values of $\lambda_2$ as 1.0 and $\lambda_1$ as 0.5 (see Eq. \ref{eq:final}) respectively, while training T-ANACTO. Also, and the weight for feature loss is set to 1.0 For training and testing, our model takes 10 sampled frames as input and takes 1s to process a batch of 4 clips during inference. We keep the required input number of frame for each baseline method as proposed in their original paper. 

We used annotations from \cite{Shan20} detector for identifying active objects in observed segment and to train the model for all datasets with $\mathcal{L}_{cao}$ loss.
Specifically, for EK-100 \cite{ek100} and EGTEA+ datasets,
during training, we maintained a lookup window of 10 frames from starting frame of action to look for first identified location of active objects \textit{i.e; bounding boxes} (if visible) to be labeled as ground truth for ANACTO task. 
It is also possible that for some clips, \emph{true contact} i.e., the actual interaction with an object can start sometime later after our lookup window. For those cases, we do not get bounding box labels for the location of active object. This means no object was actually active during the start of the action segment. However, we checked whether if this situation does lead to any inconsistency and observed that
an active object is present \emph{94\% and 92\%} of the times in the first 10 frames of the action segment for the EK-100 and EGTEA dataset, respectively. 
It is important to notice that EK-100 and EGTEA do not supply object detections. As mentioned before, to obtain this information, we rely on Faster-RCNN \cite{fastercnn} provided by \cite{ek100} pre-trained on EK-55 \cite{ek55} to detect the location of every object in the scene with a confidence score associated with each prediction \textit{b} $\in$ $\mathbb{R}^5$. 
For both datasets, we use the training and test splits provided by \cite{rulstm} for the evaluations of the T-ANACTO and the baseline methods. On the other hand, for Ego4D, we used the forecasting split for training and validation provided by \cite{ego4d}.
It is important to notice that the annotations provided for \emph{NAO} are \textit{wrt.} only \emph{the last observed frame}. As Ego4D is highly big-scaled, it was not possible to annotate it as we performed for other datasets. Therefore, we utilized only the supplied data as the ground-truth. On the other hand, this allowed us to show another utility of the ANACTO task, i.e., its setup also works for the model(s) to forecast \emph{NAO} in the last observed frame.

\section{Results}
\label{sec:results}
As the evaluation metrics, Average Precision ($AP$) with various IoU thresholds: 5, 10, 20 and 50 as well as their average shown as $AP_{avg}$ were used. \\

\noindent
\textbf{The Effect of Losses and The Backbone.} We first present an ablation study to evaluate losses given in Eq. \ref{eq:final} as well as testing a different backbone (i.e., ResNet101, notice that this is the backbone used by TSN \cite{TSN2016ECCV}) while keeping the other settings of T-ANACTO the same. The corresponding results are given in
Table \ref{table:loss}, when the experiments were performed on EK-100 dataset and the anticipation length $\tau_a = 0.25s$. As seen, using the transformer backbone compared to ResNet101 improves the results for all cases (ResNet101 vs. T-ANACTO w/ $\mathcal{L}_{nao}$ and ResNet101 vs. T-ANACTO w/ $\mathcal{L}_{cao}$+$\mathcal{L}_{nao}$). Moreoever, $\mathcal{L}_{cao}$ brings in important performance improvements to the ANACTO task, highlighting the importance of using the object-centric features.

\begin{table}[h!]
\begin{center}
\resizebox{\columnwidth}{!}{
\begin{tabular}{|l|cccc|c|} \hline
Ablation & AP5 & AP10 & AP20 & AP50 & $AP_{avg}$  \\ \hline
ResNet101 & 31.2 & 28.1 & 17.4 & 2.3 & 19.75 \\
T-ANACTO w/ $\mathcal{L}_{nao}$ & 33.5 & 29.6 & 19.3 & 2.4 & 21.2 \\
T-ANACTO w/ $\mathcal{L}_{cao}$+$\mathcal{L}_{nao}$ (FULL)& \textbf{37.1} & \textbf{32.6} & \textbf{21.1} & \textbf{4.1} & \textbf{23.7} \\ \hline
\end{tabular}}
\end{center}
\caption{Ablation study performed on EK-100 \cite{ek100} to investigate the effect of losses and the backbones of T-ANACTO.}
\label{table:loss}
\vspace{-20pt}
\end{table}

\begin{table*}[t]
\begin{center}
\resizebox{\linewidth}{!}{
\begin{tabular}{|l||cccc|c||cccc|c|cccc|c|} \hline
 Anticipation time & \multicolumn{5}{c|}{$\tau_a =$ 1.0 s} & \multicolumn{5}{c||}{$\tau_a =$ 0.5 s} & \multicolumn{5}{c|}{$\tau_a =$ 0.25 s}\\\hline\hline
 Models & AP5 & AP10 & AP20 & AP50 & $AP_{avg}$   & AP5 & AP10 & AP20 & AP50 & $AP_{avg}$  & AP5 & AP10 & AP20 & AP50 & $AP_{avg}$ \\\hline\hline
 AVT \cite{avt} & 25.2 & 19.1 & 13.6 & 1.5 & 14.9 & 30.0 & 26.4 & 17.2 & 3.1 & 19.2 & 32.3 & 27.1 & 18.4 & 3.3 & 20.2  \\ 
RULSTM \cite{rulstm} & 27.6 & 21.3 & 14.2 & 2.1 & 16.3 & 29.5 & 24.2 & 15.5 & 3.0 & 18.0 & 31.6 & 25.8 & 16.6 & 3.1 & 19.3 \\
 TSN(rgb) \cite{TSN2016ECCV} & 17.2 & 12.1 & 7.6 & 0.7 & 9.4 & 20.2 & 16.4 & 8.6 & 1.7 & 11.7 & 25.6 & 19.1 & 11.8 & 1.8 & 14.6  \\ 
 RULSTM(obj) \cite{TSN2016ECCV} & 24.4 & 19.3 & 11.1 & 1.7 & 14.1 & 24.4 & 19.1 & 11.3 & 1.8 & 14.1 & 27.0 & 20.2 & 14.7 & 1.9 & 16.0  \\
 Liu et al. \cite{liu2019forecasting} & 13.1 & 9.8 & 5.2 & 0.4 & 7.1 & 13.4 & 10.7 & 5.6 & 0.6 & 7.6 & 14.7 & 10.4 & 5.6 & 0.7 & 7.9  \\ \hline
 \textbf{T-ANACTO} & \textbf{34.4} & \textbf{28.8} & \textbf{18.1} & \textbf{3.2} & \textbf{21.2}  & \textbf{35.4} & \textbf{29.7} & \textbf{20.2} & \textbf{3.3} & \textbf{22.1}  & \textbf{37.1} & \textbf{32.6} & \textbf{21.1} & \textbf{4.1} & \textbf{23.7}  \\ 
\hline
\end{tabular}}
    
\end{center}
\caption{Results of our T-ANACTO model and other baseline methods for different TTC duration, i.e., 1, 0.5 and 0.25 seconds, tested on the EK-100 \cite{ek100}. Best result of each column are given in bold.}
\label{table:effectofAntipationLength}
\vspace{-20pt}
\end{table*}

\noindent
\paragraph{{\textbf{Effect of Anticipation Length.}}} We compare the performances of T-ANACTO and the baseline methods for various anticipation lengths for the ANACTO task in the unobserved scenes. This set of experiments was realized on EK-100 dataset \cite{ek100} and the corresponding results are given in Table \ref{table:effectofAntipationLength}. It is important to mention that since we keep the total number of sampled frames from a given observed clip as constant throughout the experiments, the change in anticipation time $\tau_a$ also changes the observed length $\tau_o$ of the clip. In other words, in these sets of experiments, the decrease in anticipation length $\tau_a$ also reduces the respective observed length $\tau_a$ of time duration. The results given in Table \ref{table:effectofAntipationLength} show that changing the anticipation lengths from higher values to lower values (e.g., from 1s to 0.5s or from 0.5s to 0.25s), as expected, increases the performance of T-ANACTO as well as all baseline methods. \\

\noindent
\textbf{Comparisons among T-ANACTO and Baselines.}
Table \ref{table:effectofAntipationLength} presents a performance comparison among T-ANACTO and baseline methods on EK-100 dataset. As seen, our method T-ANACTO surpasses all the other methods in all metrics, for all TTC durations, while the second-best method is chaining for different TTC durations. We also present comparisons on EGTEA+ and Ego4D datasets in Tables \ref{table:egt} and \ref{table:ego4d}, respectively, when the TTC duration $\tau_a$ is 0.25s for EGTEA and rate of sampling frames is 0.25s for Ego4D. To do so, for EGTEA+, we used training and testing splits-1 (see \cite{rulstm} for details) and for Ego4D, the experiments were conducted with the training and validation splits provided for the forecasting task. As mentioned in Sec. \ref{sec:dataset}, the \emph{NAO} for Ego4D is identified at the end of the past observed segment. Even for this setup, we notice that the attention-based mechanism elevated by object centric information performs better, compared to other baselines. The obtained results in the aforementioned tables are in line with the results obtained for the EK-100 dataset, showing that T-ANACTO outperforms the other baseline methods, while the performance improvement can be up to 12\% in terms of $AP_{avg}$.

\begin{table}[t]
\begin{center}
\resizebox{\linewidth}{!}{
\begin{tabular}{|l|cccc|c|} \hline
Models & AP5 & AP10 & AP20 & AP50 & $AP_{avg}$  \\\hline\hline
AVT \cite{avt} & 19.7 & 16.5 & 10.2 & 2.6 & 12.2 \\ 
RULSTM \cite{rulstm} & 18.8 & 13.4 & 7.7 & 1.4 & 10.3 \\
TSN(rgb) \cite{TSN2016ECCV} & 14.8 & 12.1 & 7.4 & 1.4 & 9.0 \\ 
RULSTM(obj) \cite{rulstm} & 15.1 & 12.4 & 6.8 & 1.3 & 9.0 \\
Liu et al. \cite{liu2019forecasting} & 11.8 & 8.5 & 5.7 & 1.0 & 6.8 \\
\textbf{T-ANACTO} & \textbf{26.6} & \textbf{21.0} & \textbf{14.7} & \textbf{2.8} & \textbf{16.3} \\ \hline
\end{tabular}}
\end{center}
\caption{T-ANACTO and the baseline methods' performances when they are tested on EGTEA+ dataset \cite{egtea} with the TTC duration $\tau_a = 0.25s$. Best result of each column are given in bold.}
\label{table:egt}
\vspace{-10pt}
\end{table}

 \begin{table}[t]
\begin{center}
  \resizebox{\linewidth}{!}{
 \begin{tabular}{|l|cccc|c|} \hline
 Models & AP5 & AP10 & AP20 & AP50 & $AP_{avg}$  \\\hline\hline
 AVT \cite{avt} & 38.8 & 28.7 & 12.9 & 2.9 & 20.8 \\
 RULSTM \cite{rulstm} & 37.6 & 27.4 & 10.3 & 1.7 & 19.3 \\
 TSN(rgb) \cite{TSN2016ECCV} & 35.5 & 23.2 & 8.5 & 1.5 & 17.1 \\
 RULSTM(obj) \cite{rulstm} & 34.6 & 21.3 & 8.2 & 1.5 & 16.4 \\
 Liu et al. \cite{liu2019forecasting} & 15.2 & 11.1 & 7.4 & 1.1 & 8.7 \\
 \textbf{T-ANACTO} & \textbf{41.2} & \textbf{31.4} & \textbf{18.6} & \textbf{4.6} & \textbf{24.0} \\ \hline
 \end{tabular}
}
\end{center}
\caption{T-ANACTO and the baseline methods' performances when they are tested on Ego4D dataset \cite{ego4d} to identify NAO. Frames are sampled from observed segment at $0.25s$. Best result of each column are given in bold.
}
 \label{table:ego4d}
 \vspace{-20pt}
 \end{table}

\paragraph{\textbf{Qualitative Results.}}

\begin{figure}[t]
\includegraphics[width=\columnwidth]{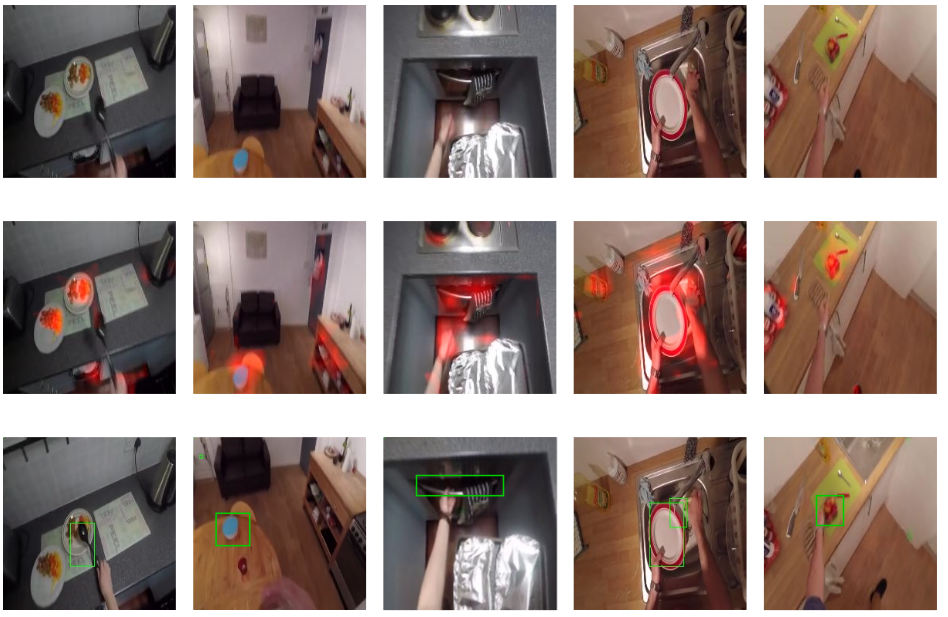}
\caption{The top row shows the ``last observed frame", the middle row shows ``the region of interest of T-ANACTO", and the bottom row shows ``the starting frame of an action". The green box(es) in the last row represent the location of \emph{NAO} bounding box in the starting frame(s) of action.}
\vspace{-0.5cm}
\label{fig:qualitative}
\end{figure}

We visualize the effective spatial attention by our T-ANACTO encoder on the last observed frame in Fig. \ref{fig:qualitative} for EK-100 \cite{ek100}. The red regions demonstrate the regions of interest to the model, which correspond to human-object interaction in the future frames and help in anticipating the \emph{NAO}.  The results show that our model learns to focus on objects which are likely to be in contact with human hands based on observation till last observed frame, and thus the inference can also be performed before the contact happens. Notice in the second column, even though the object is not active in the starting frame, our model learns to focus on a possible object which becomes active later. We also notice that the model performs equally well for different lighting conditions. Besides, it is also interesting to note that T-ANACTO model is also able to identify human-interaction hotspots for an object in some case. 
In the Supp. Material, we provide more qualitative results of our model for identifying objects in last observed frame, different TTC $\tau_a$, and discuss failure cases for the model.

\section{Conclusions}
We have investigated the problem of anticipating next active object localization. First, we discussed the formulation of the ANACTO task. We then presented a new vision transformer based model, T-ANACTO which learns to encode hand-object interactions with the help of an object detector. We proved its effectiveness by comparing it against relevant strong anticipation based baseline methods. The experimental evaluation highlights that: (1) the object-centered cues help in elevating the performance to locate the next possible active object; (2) the effectiveness of the model increases when the anticipation time for the prediction before the beginning of an action is kept short. Besides, we also discuss the effect of observation length on the performance of model(s). (3) Our model effectively learns to identify and allocate attention to possible action objects in the future, as realized from qualitative results. (4) Importantly, T-ANACTO is also able to detect \emph{NAO} location even in the last observed frame. Finally, we also supply the ANACTO task annotations for EGTEA+ and EK-100 datasets, i.e., hand and active object bounding box annotations along with their contact state as well as providing the object annotations for the entire dataset using an object detector pre-trained on EK-55 \cite{ek55}. 

As future work, we will extend the ANACTO task to predict the dynamic TTC, noun and verb for \emph{NAO}, and investigate the use of an object tracker with other human-centered cues such as gaze and the appearance of objects over time. We will also investigate the effect of action recognition on \emph{NAO} identification and localization.

\clearpage


{\centering\textbf{Anticipating Next Active Objects for Egocentric Videos: Supplementary Material}} 
\vspace{30pt}

\maketitle

This supplementary material includes the visualization of the attention maps of our T-ANACTO encoder, which is given for different time to contact window (Section \ref{sec:visualizationAttention}) for EpicKitchen \cite{ek100, ek55} and EGTEA \cite{egtea} datasets. The attention maps provide an intuition on learning of T-ANACTO to identify ``interactable'' objects (i.e; possible next-active-object) in the scene and then model the motion of the person till its TTC to anticipate its contact location in future frame. We also provide further visualizations for Ego4D \cite{ego4d} dataset when the next-active-object is identified at the last observed frame irrespective of time to contact with that object. Then, we discuss the failure cases of our model in Section \ref{sec:cases} through multiple exemplary images. In addition, we also provide \textbf{a video} giving the details of the transition of attention over past frames till the last observed frame in a video clip.

\section{Visualization of the Attention Maps}
\label{sec:visualizationAttention}
For training purposes, the vision transformer (VIT) \cite{vit} model was implemented using the timm \cite{timm} based pytorch-image model, which does not provide attention weights for the output of transformer encoder. To 
visualize the spatial attention of our T-ANACTO encoder, we implemented a similar model 
with the same nomenclature for the layers to load the trained weights from the training of the model. The attention weights are then extracted from each block layer and then stacked together to project the learning of our encoder.

\begin{figure}[t]\centering
\includegraphics[width=1.0\columnwidth]{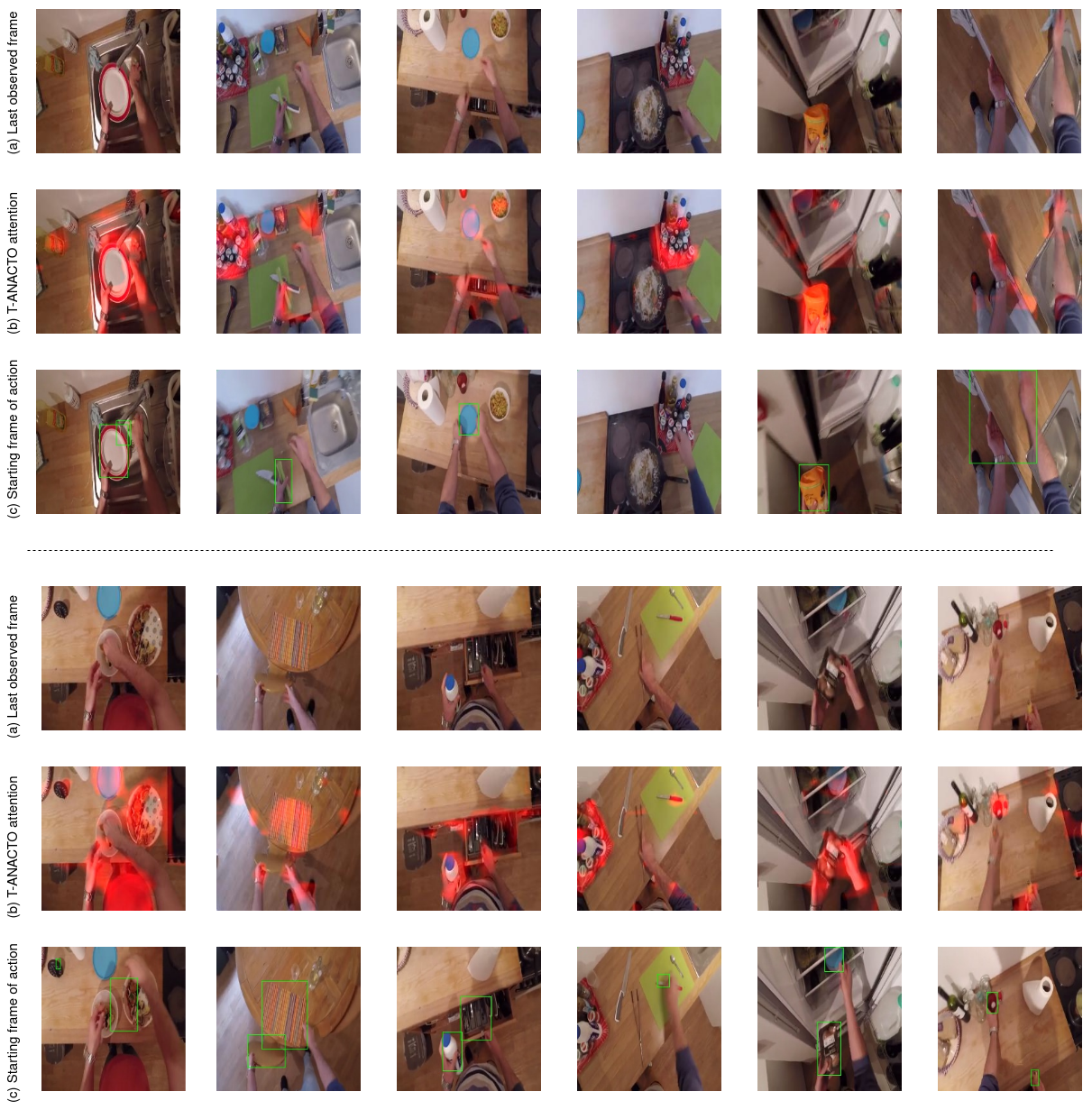}
\caption{Results showing the attention map generated by our T-ANACTO encoder for last observed frame of video clip with TTC $\boldsymbol{\tau_a = 0.25}$ seconds before the beginning of the action. The red regions depicts the region of interest to identify the next active object in the starting frame of the action. The green bounding box for the starting frame of the action (row) shows the localization of the active object for that frame. It is interesting to note that for segments which there is no active object at the start of the action, our encoder is able to identify the possible area of interest for next future frames post the starting frame of the action. }
\label{fig:ek_25}
\end{figure}

\begin{figure}[t]\centering
\includegraphics[width=0.9\columnwidth]{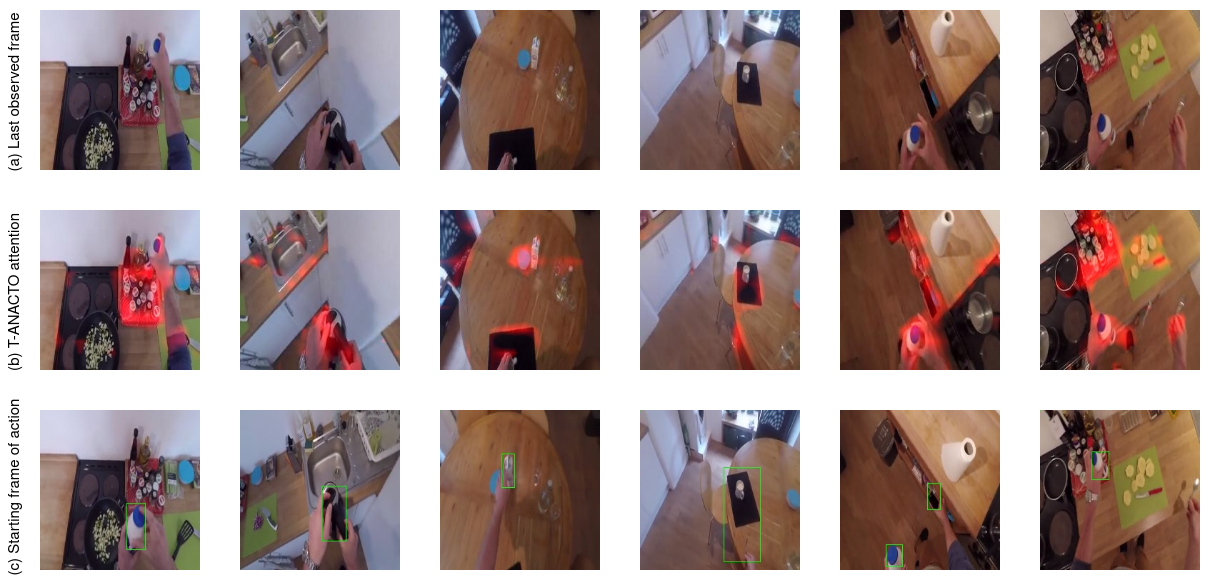}
\caption{Results showing the attention map generated by our T-ANACTO encoder for last observed frame of video clip with TTC $\boldsymbol{\tau_a = 0.5}$ seconds before the beginning of the action. }
\label{fig:ek_5}
\end{figure}

\begin{figure}[t]\centering
\includegraphics[width=0.9\columnwidth]{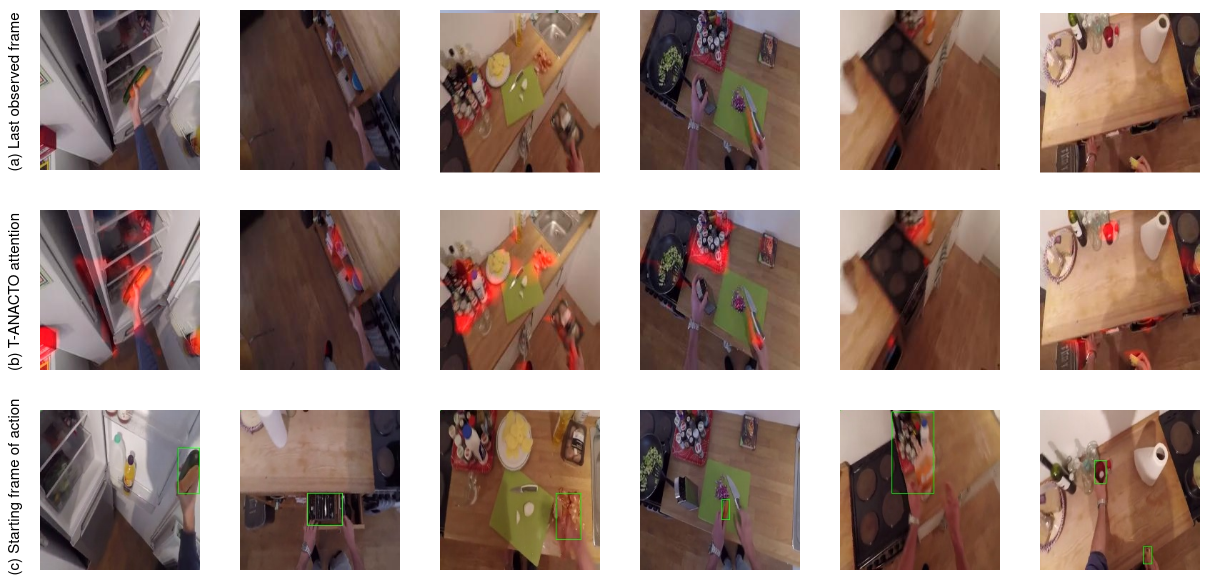}
\caption{Results showing the attention map generated by our T-ANACTO encoder for last observed frame of video clip with TTC $\boldsymbol{\tau_a = 1.0}$ second before the beginning of the action. }
\label{fig:ek_1}
\end{figure}

\begin{figure}[t]\centering
\includegraphics[width=0.9\columnwidth]{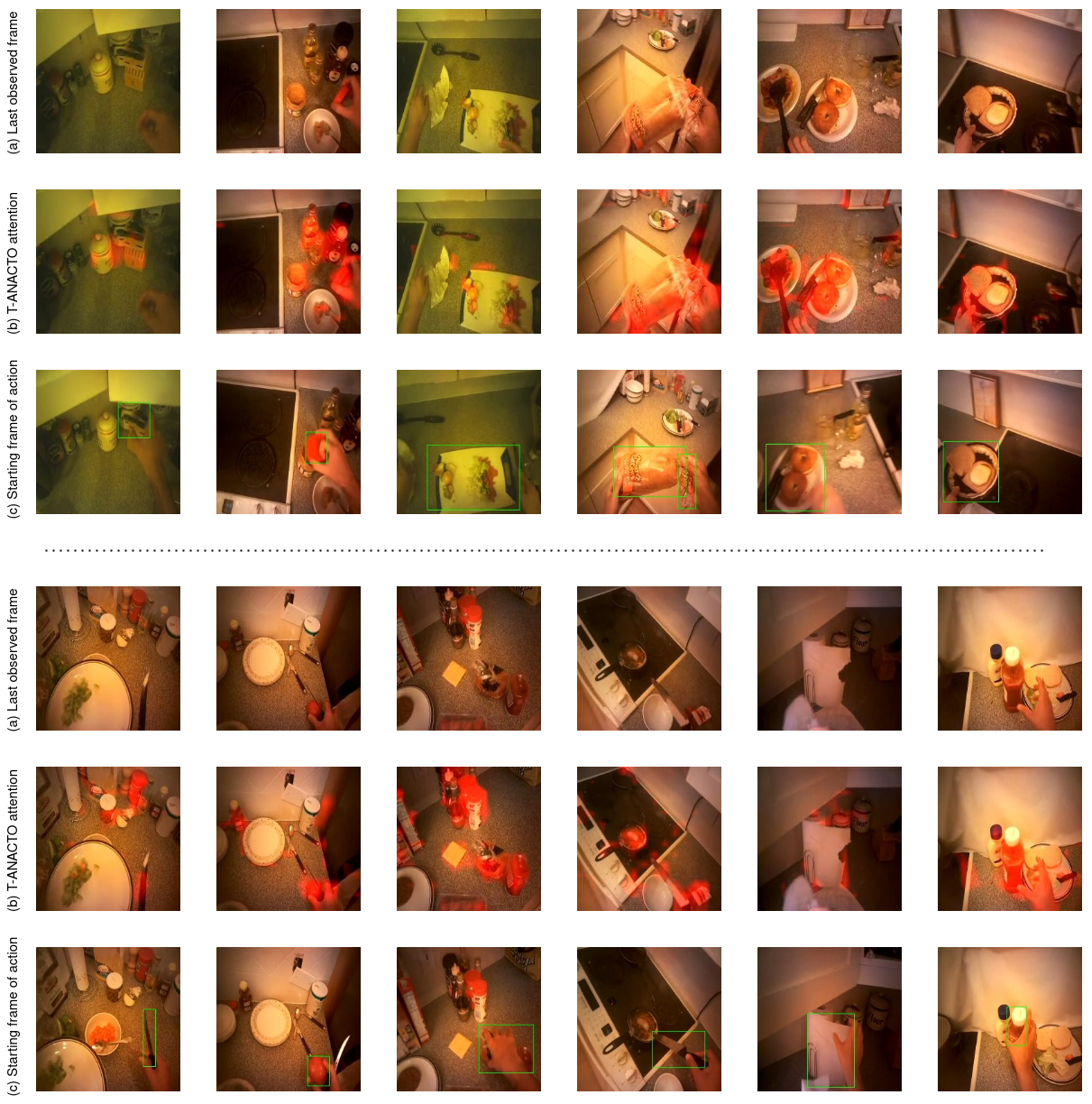}
\caption{Results shows the spatial attention map for EGTEA+ dataset. The green bounding box specifies the location of the active object at the starting of an action. }
\label{fig:egtea}
\end{figure}

We show the effectiveness of our model T-ANACTO for anticipating next active object task (ANACTO) as spatial attention of our encoder in additional figures for both   EpicKitchen-100 \cite{ek100} and EGTEA+ \cite{egtea} datasets in Fig. \ref{fig:ek_25}, \ref{fig:ek_5}, \ref{fig:ek_1} and \ref{fig:egtea}. Using this visualization, one can understand how the confidence of the model differs as it analyzes frames that are temporally distant from the beginning of an action segment for different TTC window $\tau_a$. 
In other words, we are able to compare the diversity of attention for different TTC window, $\tau_a$ \textit{v/s} observed $\tau_o$ time of video clips for EpicKitchen \cite{ek100} dataset.

In detail, extending on the visualization provided in the main paper for EpicKitchen-100 dataset, herein, we report additional results for that dataset for different TTC window $\tau_a =$ 0.25 seconds, 0.5 seconds, 1.0 second in Fig. \ref{fig:ek_25}, \ref{fig:ek_5}, \ref{fig:ek_1}, respectively. 
In the mentioned figures, we report the last observed frame by the model and the attention map generated for that particular frame by our T-ANACTO encoder to predict the location of next active object. We also report the ground truth results for the active object at the starting frame of action segment. This attention map(s) is generated after considering the past frames and the last frame of observed segment $\tau_o$. 
As mentioned in the main paper, the change in $\tau_a$ for a video clip also affects the observed video segment length $\tau_o$ proportionally.

\begin{figure}[t]\centering
\includegraphics[width=0.9\columnwidth]{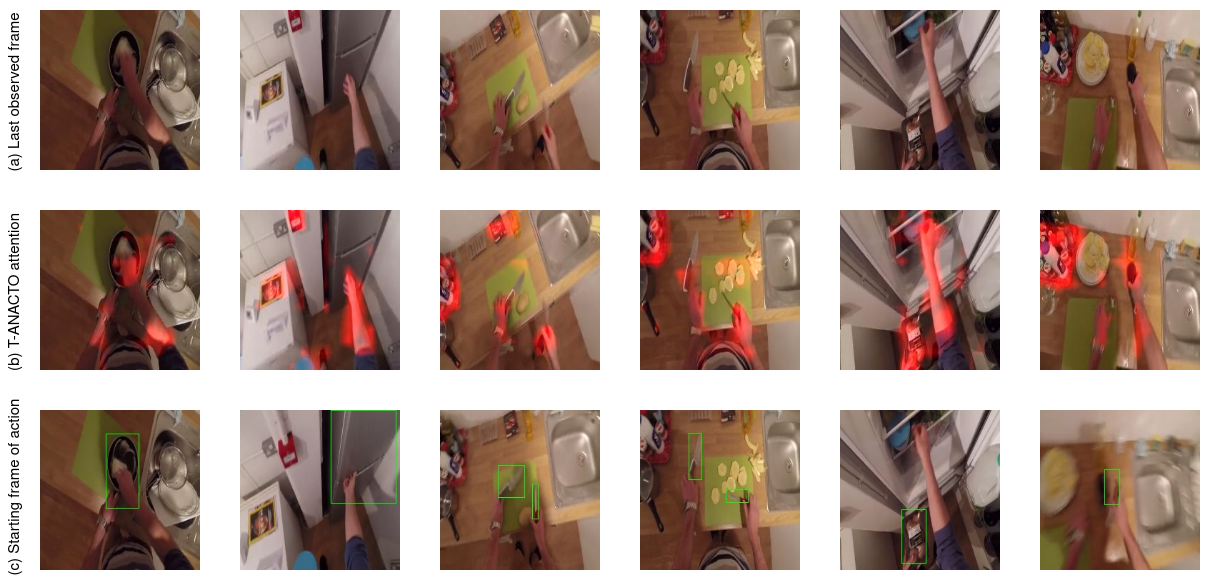}
\caption{Our T-ANACTO model is designed to identify and locate the hand-object interaction location for future frames. In the process, it also learns to attribute to hand position in an image frame without explicitly providing hand location. The figures provided illustrate the spatial attention of our model to hand positions besides possible next active object.}
\label{fig:detections}
\end{figure}

\begin{figure}[t]\centering
\includegraphics[width=0.9\linewidth]{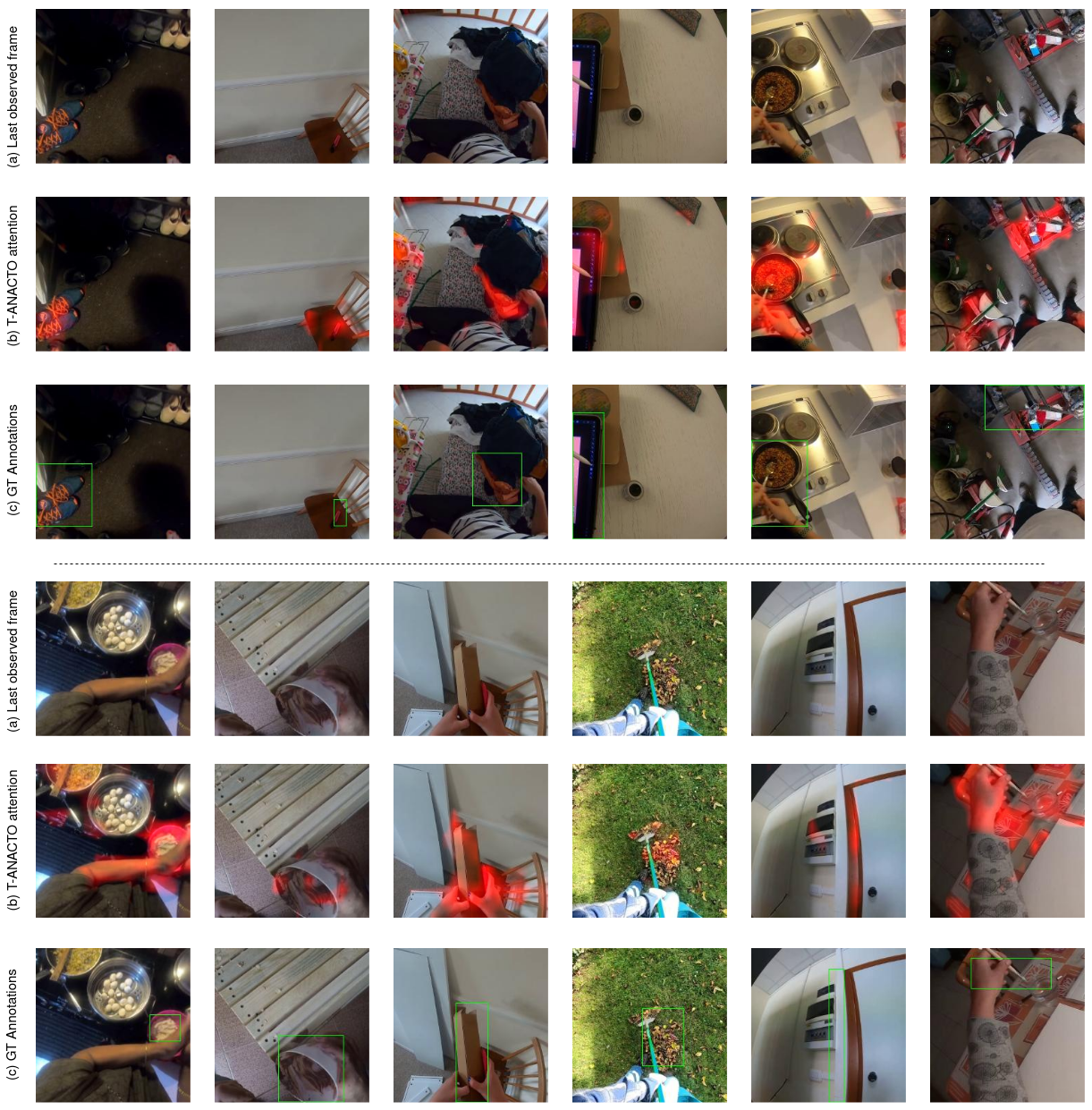}
\caption{Results shows the spatial attention map for ego4d dataset when trained to identify next active object \textit{wrt} last observed frame. The green bounding box specifies the location of the object which will become active in the future. The highlighted region specifies the attention stress by the model in the last observed frame. }
\label{fig:ego4d}
\end{figure}

\begin{figure}[t]\centering
\includegraphics[width=0.9\columnwidth]{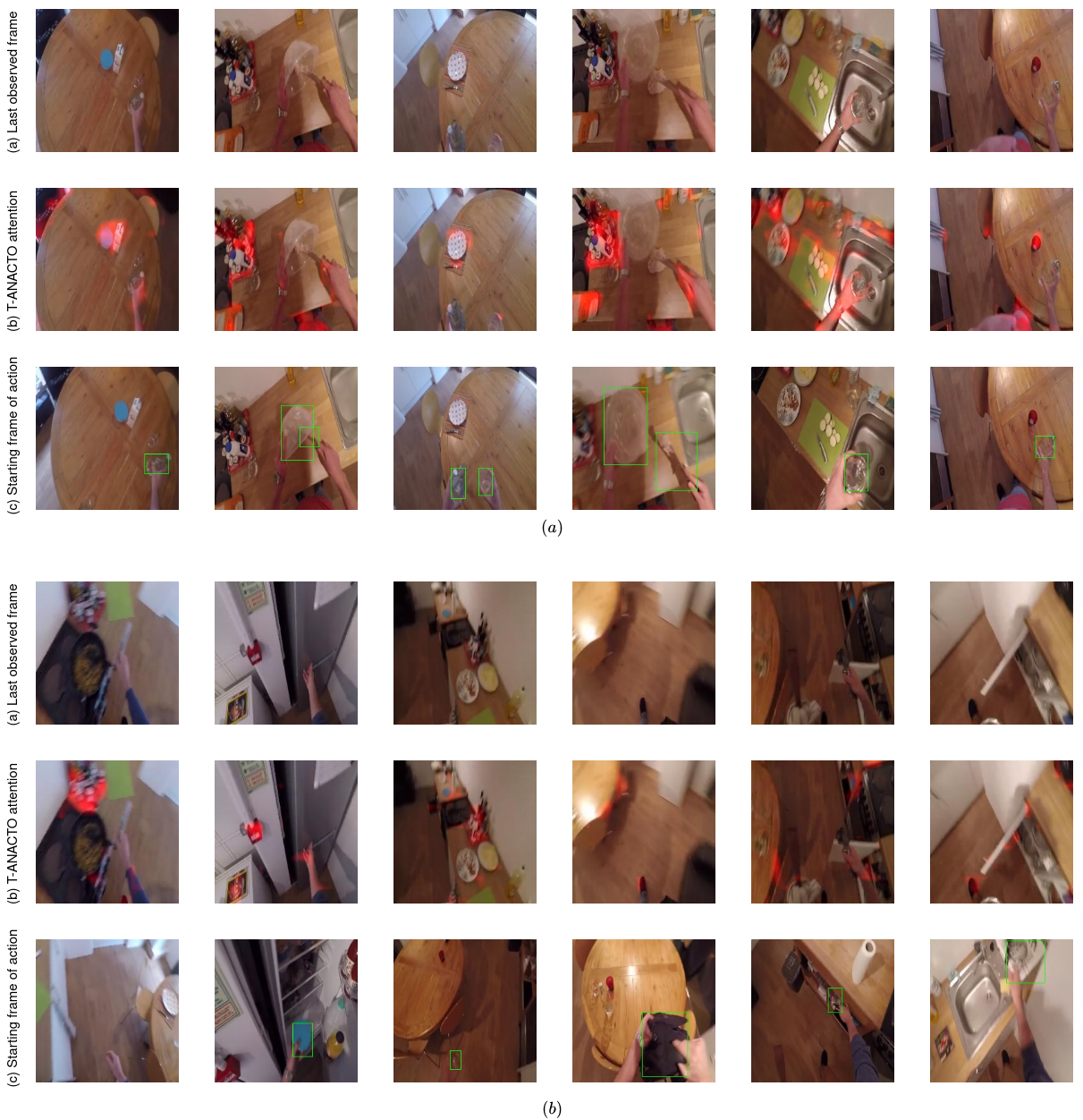}
\caption{In the figure, we report results for some of the failure cases of our model as discussed in Section \ref{sec:fail_ek_gtea}. (a) The model fails to attribute attention to objects which are light colored or easily camouflaged with the background. 
(b) When the scene completely changes at the beginning of action from the past observed segment.}
\label{fig:fail}
\end{figure}

\begin{figure}[t]\centering
\includegraphics[width=0.9\columnwidth]{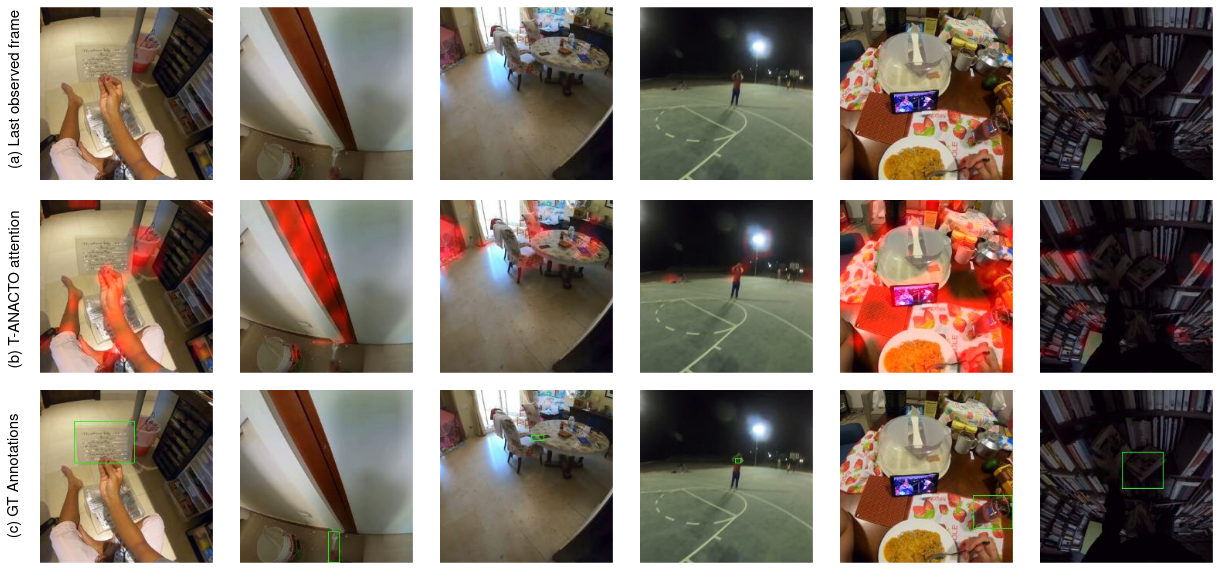}
\caption{In the figure, we report results for some of the failure cases of our model as discussed in Section \ref{sec:fail_ego} for Ego4D dataset. (a) The model fails to attribute for higher TTC time for a given next active object. 
(b) For objects which are tiny or transparent or scattered around multiple objects it is difficult to identify next active object for larger TTC.}
\label{fig:fail_ego}
\end{figure}

\begin{figure*}[t]\centering
\includegraphics[width=0.9\linewidth]{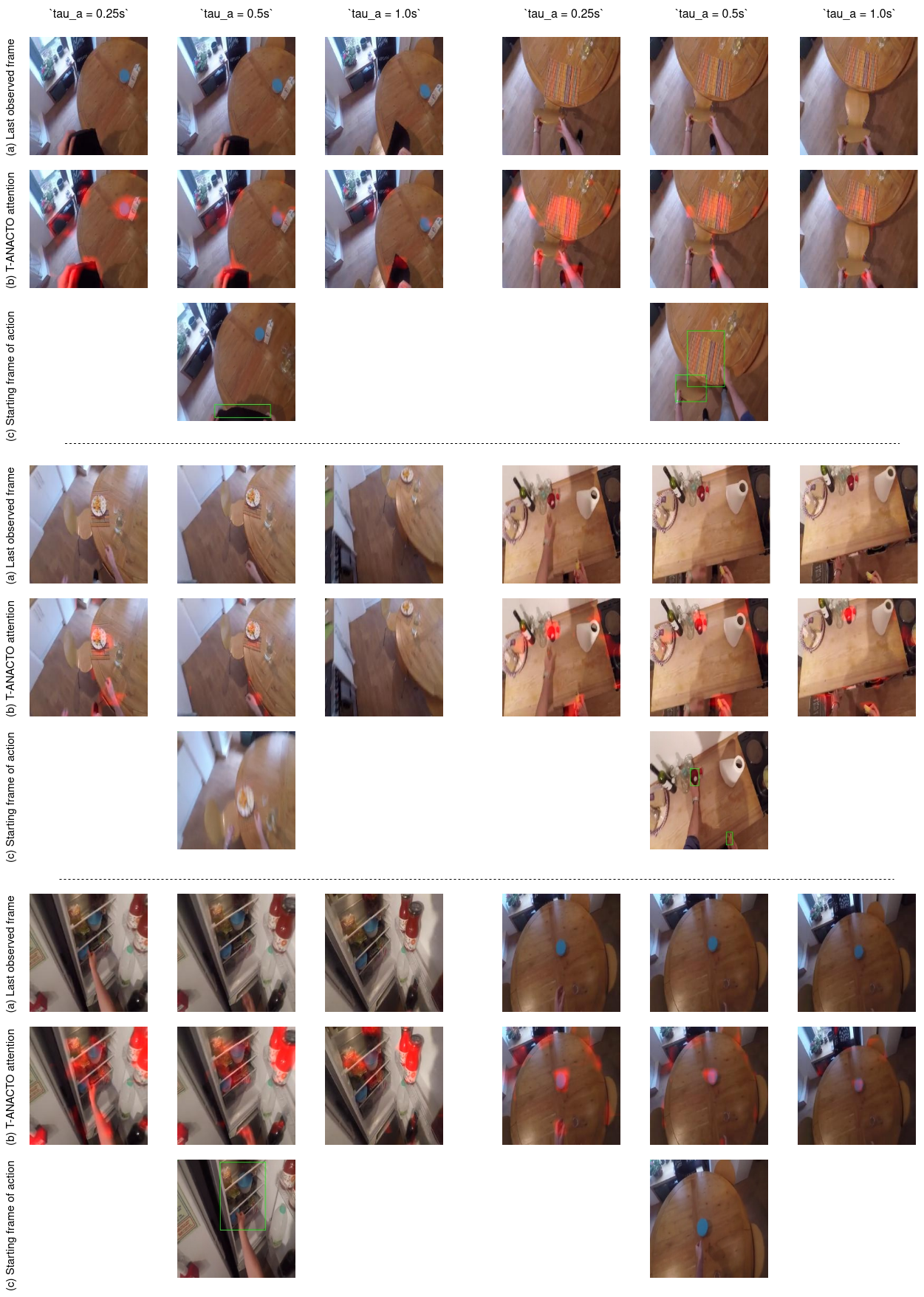}
\caption{Results show the diversity of spatial attention for the last observed frame preceding the beginning of an action segment for different setups of TTC window for $\tau_a = 0.25, 0.5, 1.0$ seconds. The attention corresponds to the red region in the image. The regions tend to appear more assertive as the model examines frame closer to action segments \textit{i.e;} as the $\tau_a$ is decreased. This also attributes to a higher accuracy of the model for shorter time to contact window.}
\label{fig:compare}
\end{figure*}

To qualitatively understand the improved performance of the model as the $\tau_a$ is reduced from $1.0s$ to $0.25$ seconds, we report the comparison in Fig. \ref{fig:compare}. It is visible that as the model is fed with frames that are closer to the beginning of an action segment, \textit{i.e lower $\tau_a$}, its confidence for the next active object increases and so the performance gain can be justified.

It is also important to mention that, for most of the results, one can notice that our model is also able to identify the hand's positions and interaction hotspots for certain objects, although our model does not explicitly require the hand's position as an input. We confirm this by reporting our results for the EpicKitchen-100 dataset \cite{egtea} in Fig. \ref{fig:detections}. Since our method learns to identify the future hand-object interaction it focuses on locating the position of hands and respectively locate the next active object in consequent starting frame of an action segment.

In Fig. \ref{fig:egtea}, we provide the attention map for the learning of our model for EGTEA+ dataset \cite{egtea} when trained on \textit{train split 1} and tested on \textit{test split 1}. 
We also provide \textbf{a video} visualizing the transition of attention on all past frames till the last observed frame for different clips. 

\textbf{Ego4D \cite{ego4d}.} In Fig. \ref{fig:ego4d}, we provide the attention map for the learning of our model for Ego4D dataset when trained on training set of forecasting split to predict the next active object location \textit{wrt} last observed frame.

\section{Success and Failure cases}
\label{sec:cases}
All the visualizations discussed in the previous section, is given for the cases T-ANACTO is successful to anticipate the next active object correctly. 
In this section, we discuss the cases which can be considered as failure for T-ANACTO.

\subsection{Epic Kitchen and EGTEA}
\label{sec:fail_ek_gtea}
We were able to identify two major cases for EpicKitchen and EGTEA: 

\paragraph{\textbf{1) Light colored objects.}} We noticed that the model is not able to confine its attention to those areas in the video clips  where a light colored or transparent object is used for human-object interaction (see Fig. \ref{fig:fail}(a)). 
It could perhaps be a failure of the object detection model which is not able to identify items due to the transparent nature of object and camouflage with the background of frames(s). However, for most of the video clips consisting of light colored objects, our model is able to identify the hand's positioning in the frames as described in Fig. \ref{fig:detections} which can be exploited to further extend the work in this domain. 
\vspace{-0.5cm}

\paragraph{\textbf{2) Scene transition.}} As stated earlier, the next active object detection is a challenging task due to the consistent nature of humans to continuously interact with the environment. In the process, a person does the interaction with the objects based on the activities being performed which can lead to sudden change of scenes from one moment to another. 
Therefore, a current scene at the start of action segment might be drastically different \textit{wrt} past observed frames. In those cases, it is extremely difficult for the model to locate "interactable" objects in the scene which has not be observed by the model  (see Fig. \ref{fig:fail}(b)). 

\subsection{Ego4D} We provide the visualization of cases in Fig. \ref{fig:fail_ego}
\label{sec:fail_ego}
\paragraph{\textbf{1) Sampling of frames.}} Since our model take input frames at a sampled interval, it is trained to output predictions after the the sampled interval time after the last observed frame. However, in Ego4D dataset the TTC for a next active object varies drastically for each clip, which is one of the main reasons our model suffers for those objects whose TTC are much higher than sampled frame rate for our input frames. 
\vspace{-0.5cm}
\paragraph{\textbf{2) Tiny and clustered objects.}} We also notice that our model fails for tiny / transparent objects in the scene or where multiple objects are scattered in the frame.


\clearpage

{\small
\bibliographystyle{ieee_fullname}
\bibliography{main}
}

\end{document}